\documentclass[11pt,a4paper]{article}
\usepackage[hyperref]{emnlp2020}
\usepackage{times}
\usepackage{amsmath}
\usepackage{amssymb}
\usepackage{latexsym}

\usepackage{graphicx}

\usepackage{microtype}

\aclfinalcopy 

\title{Conditional Neural Generation using Sub-Aspect Functions \\ for Extractive News Summarization}

\author{Zhengyuan Liu, \ Ke Shi, \ Nancy F. Chen \\
  Institute for Infocomm Research, A*STAR, Singapore \\
  \texttt{\{liu\_zhengyuan,shi\_ke,nfychen\}@i2r.a-star.edu.sg}}

\date{}

\begin{document}
\maketitle

\begin{abstract}
Much progress has been made in text summarization, fueled by neural architectures using large-scale training corpora. However, in the news domain, neural models easily overfit by leveraging position-related features due to the prevalence of the inverted pyramid writing style. In addition, there is an unmet need to generate a variety of summaries for different users. In this paper, we propose a neural framework that can flexibly control summary generation by introducing a set of sub-aspect functions (i.e. importance, diversity, position). These sub-aspect functions are regulated by a set of control codes to decide which sub-aspect to focus on during summary generation. We demonstrate that extracted summaries with minimal position bias is comparable with those generated by standard models that take advantage of position preference. We also show that news summaries generated with a focus on diversity can be more preferred by human raters. These results suggest that a more flexible neural summarization framework providing more control options could be desirable in tailoring to different user preferences, which is useful since it is often impractical to articulate such preferences for different applications \textit{a priori}.
\end{abstract}

\section{Introduction}
Text summarization targets to automatically generate a shorter version of the source content while retaining the most important information. As a straightforward and effective method, extractive summarization creates a summary by selecting and subsequently concatenating the most salient semantic units in a document. Recently, neural approaches, often trained in an end-to-end manner, have achieved favorable improvements on various large-scale benchmarks \cite{nallapati2017summarunner, narayan-2018-ExtremeSumm, liu-lapata-2019-PreSumm}.

Despite renewed interest and avid development in extractive summarization, there are still long-standing, unresolved challenges. One major problem is position bias, which is especially common in the news domain, where the majority of research in summarization is studied. In many news articles, sentences appearing earlier tend to be more important for summarization tasks \cite{hong-2014-positionBias}, and this preference is reflected in reference summaries of public datasets. However, while this tendency is common due to the classic textbook writing style of the ``inverted pyramid'' \cite{christopher-1999-writingBasics}, news articles can be presented in various ways. Other journalism writing styles include anecdotal lead, question-and-answer format, and chronological organization \cite{stovall1985writingMassMedia}. Therefore, salient information could also be scattered across the entire article, instead of being concentrated in the first few sentences, depending on the chosen writing style of the journalist. 

As the ``inverted pyramid'' style is widespread in news articles \cite{kryscinski-2019-criticalSumm}, neural models would easily overfit on position-related features in extractive summarization tasks because of the data-driven learning setup which tags on to features that correlate the most with the output. As a result, those models would select the sentences at the very beginning of a document as best candidates regardless of considering the full context, resulting in sub-optimal models with fancy neural architectures that do not generalize well to other domains \cite{kedzie-2018-contentSelection}. 

Additionally, according to \citet{Nenkova-2007-human-variation}: \textit{``Content selection is not a deterministic process \cite{salton-1997-automatic-Summ,marcu-1997-discourse, mani-1997-discourse}. Different people choose different sentences to include in a summary, and even the same person can select different sentences at different times \cite{rath1961formation}. Such observations lead to concerns about the advisability of using a single human model ...''}, such observations suggest that individuals differ on what she considers key information under different circumstances. This reflects the need to generate application-specific summaries, which is challenging without establishing appropriate expectations and knowledge of targeted readers prior to model development and ground-truth construction. However, publicly available datasets only provide one associated reference summary to a document. Without any explicit instructions and targeted applications or user preferences, ground-truth construction for summarization becomes an under-constrained assignment \cite{kryscinski-2019-criticalSumm}. Therefore, it is challenging for end-to-end models to generate alternative summaries without proper anchoring from reference summaries, making it harder for such models to reach their full potential.

\begin{figure}[t]
\centering
\includegraphics[width=7.5cm]{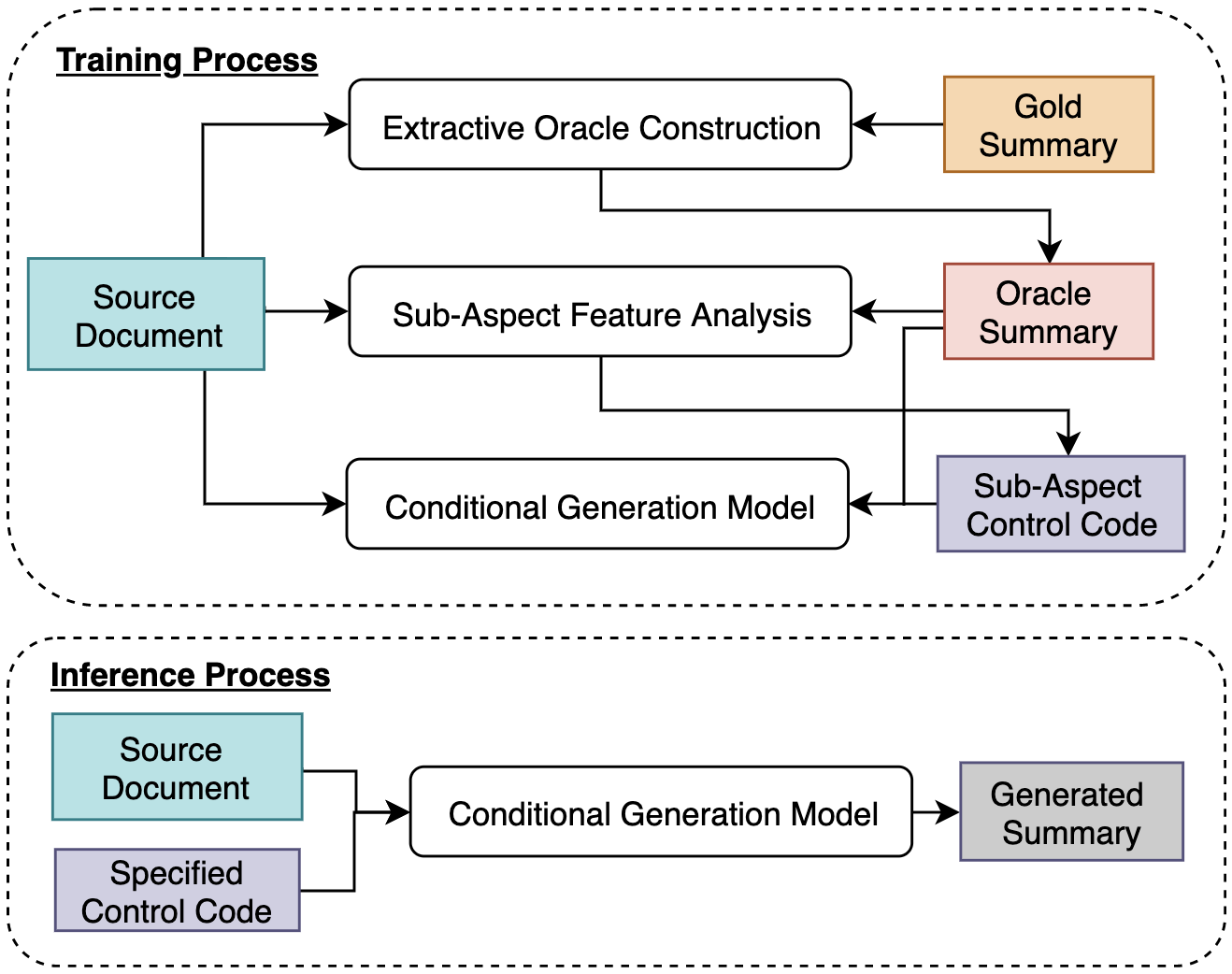}
\caption{Proposed conditional generation framework exploiting sub-aspect functions.}
\label{ctrl-system-fig}
\vspace{-0.3cm}
\end{figure}

In this work, we propose a flexible neural summarization framework that is able to provide more explicit control options when automatically generating summaries (see Figure \ref{ctrl-system-fig}). Since summarization has been regarded as a combination of sub-aspect functions (e.g. information, layout) \cite{Carbonell-1998-MMR-Summ, Lin-2012-summAspects}, we follow the spirit of sub-aspect theory and adopt control codes on sub-aspects to condition summary generation. The advantages are two-fold: (1) It provides a systematic approach to investigate and analyze how one might minimize position bias in extractive news summarization in neural modeling. Most, if not all, previous work like \cite{jung-etal-2019-earlier, kryscinski-2019-criticalSumm} only focus on analyzing the degree and prevalence of position bias. In this work, we take one step further to propose a research methodology direction to disentangle position bias from important and non-redundant summary content. 
(2) Text summarization needs are often domain or application specific, and difficult to articulate \textit{a priori} what the user-preferences are, thus requiring potential iterations to adapt and refine. However, human ground-truth construction for summarization is time-consuming and labor-intensive. Therefore, a more flexible summary generation framework could minimize manual labor and generate useful summaries more efficiently. 

An ideal set of sub-aspect control codes should characterize different aspects of summarization well in a comprehensive manner but at the same time delineate a relatively clear boundary between one another to minimize the set size \cite{higgins-2017-betaVAE}. To achieve this, we adopt the sub-aspects defined in \cite{jung-etal-2019-earlier}: \texttt{\small IMPORTANCE}, \texttt{\small DIVERSITY}, and \texttt{\small POSITION}, and assess their characterization capability on the CNN/Daily Mail news corpus \cite{hermann-2015-cnnDaily} via quantitative analyses and unsupervised clustering. We utilize control codes based on these three sub-aspect functions to label the training data and implement our conditional generation approach with a neural selector model. Empirical results show that given different control codes, the model can generate output summaries of alternative styles while maintaining performance comparable to the state-of-the-art model; modulation with semantic sub-aspects can reduce systemic bias learned on a news corpus and improve potential generality across domains.

\section{In Relation to Other Work}
In text summarization, most benchmark datasets focus on the news domain, such as NYT \cite{sandhaus-2008-NYT} and CNN/Daily Mail \cite{hermann-2015-cnnDaily}, where the human-written summaries are used in both abstractive and extractive paradigms \cite{gehrmann-2018-bottomUp}. 
To improve the performance of extractive summarization, non-neural approaches explore various linguistic and statistical features such as lexical characteristics \cite{lexical-summ-1995}, latent topic information \cite{2009-lda-summ}, discourse analysis \cite{rst-tree-summ-2015, liu-chen-2019-exploiting}, and graph-based modeling \cite{erkan-2004-lexrank, mihalcea-tarau-2004-textrank} .
In contrast, neural approaches learn the features in a data-driven manner. Based on recurrent neural networks, SummaRuNNer is one of the earliest neural models \cite{nallapati2017summarunner}. Much development in extractive summarization has been made via reinforcement learning \cite{narayan-etal-2018-RLsumm}, jointly learning of scoring and ranking \cite{zhou-etal-2018-jointSumm}, and deep contextual language models \cite{liu-lapata-2019-PreSumm}. 

Despite much development in recent neural approaches, there are still challenges such as corpus bias resulting from the prevalent ``inverted pyramid'' journalism writing style \cite{lin-hovy-1997-identifying}, and system bias \cite{jung-etal-2019-earlier} stemming from position preference in the ground-truth. However, to date only analysis work has been done to characterize the position-bias problem and its ramifications, such as inability to generalize across corpora or domains \cite{kedzie-2018-contentSelection, kryscinski-2019-criticalSumm}. Few, if any, have attempted to resolve this long-standing problem of position bias using neural approaches. In this work, we take a first stab to introduce sub-aspect functions for conditional extractive summarization. We explore the possibility of disentangling the three sub-aspects that are commonly used to characterize summarization: \texttt{\small POSITION} for choosing sentences by their position, \texttt{\small IMPORTANCE} for choosing relevant and repeating content across the document, and \texttt{\small DIVERSITY} for ensuring minimal redundancy between summary sentences \cite{jung-etal-2019-earlier} during the summary generation process. In particular, we use these three sub-aspects as control codes for conditional training. To the best of our knowledge, this is the first work in applying auxiliary conditional codes for extractive summary generation. 

In other NLP tasks, topic information is used as conditional signals and applied to dialogue response generation \cite{xing-2017-topicDialogue} and pre-training of large-scale language models \cite{keskar-2019-ctrl} while sentiment polarity is used in text style transfer \cite{john-2019-disentangledTrans}. In image style transfer, codes specifying color or texture are used to train conditional generative models \cite{mirza-2014-ConditionalGAN, higgins-2017-betaVAE}. 

\section{Extractive Oracle Construction}
\label{sec:oracle-annotation}
\subsection{Similarity Metric: Semantic Affinity vs. Lexical Overlap}
For benchmark corpora that are widely adopted, e.g. CNN/Daily Mail \cite{hermann-2015-cnnDaily}, there are only golden abstractive summaries written by humans with no corresponding extractive oracle summaries. To convert the human-written abstracts to extractive oracle summaries, most previous work used ROUGE score \cite{lin-2004-ROUGE}, which counts contiguous n-gram overlap, as the similarity criteria to rank and select sentences from the source content. 
Since ROUGE scores only conduct lexical matching using word overlapping algorithms, salient sentences from the source content paraphrased by human-editors could be overlooked as the ROUGE scores would be low, while sentences with a high count of common words could get an inflated ROUGE score \cite{kryscinski-2019-criticalSumm}. 

To tackle this drawback of ROUGE, we propose to apply the semantic similarity metric BertScore \cite{zhang-2019-bertscore} to rank the candidate sentences. BertScore has performed better than ROUGE and BLEU in sentence-level semantic similarity assessment \cite{zhang-2019-bertscore}. 
Moreover, BertScore includes recall measures between reference and candidate sequences, a more suitable metric than distance-based similarity measures  \cite{wieting-2019-SimpleSimilar,reimers-2019-senBert} for summarization related tasks, where there is an asymmetrical relationship between the reference and the generated text.

\begin{figure}[t]
\centering
\includegraphics[width=7.5cm]{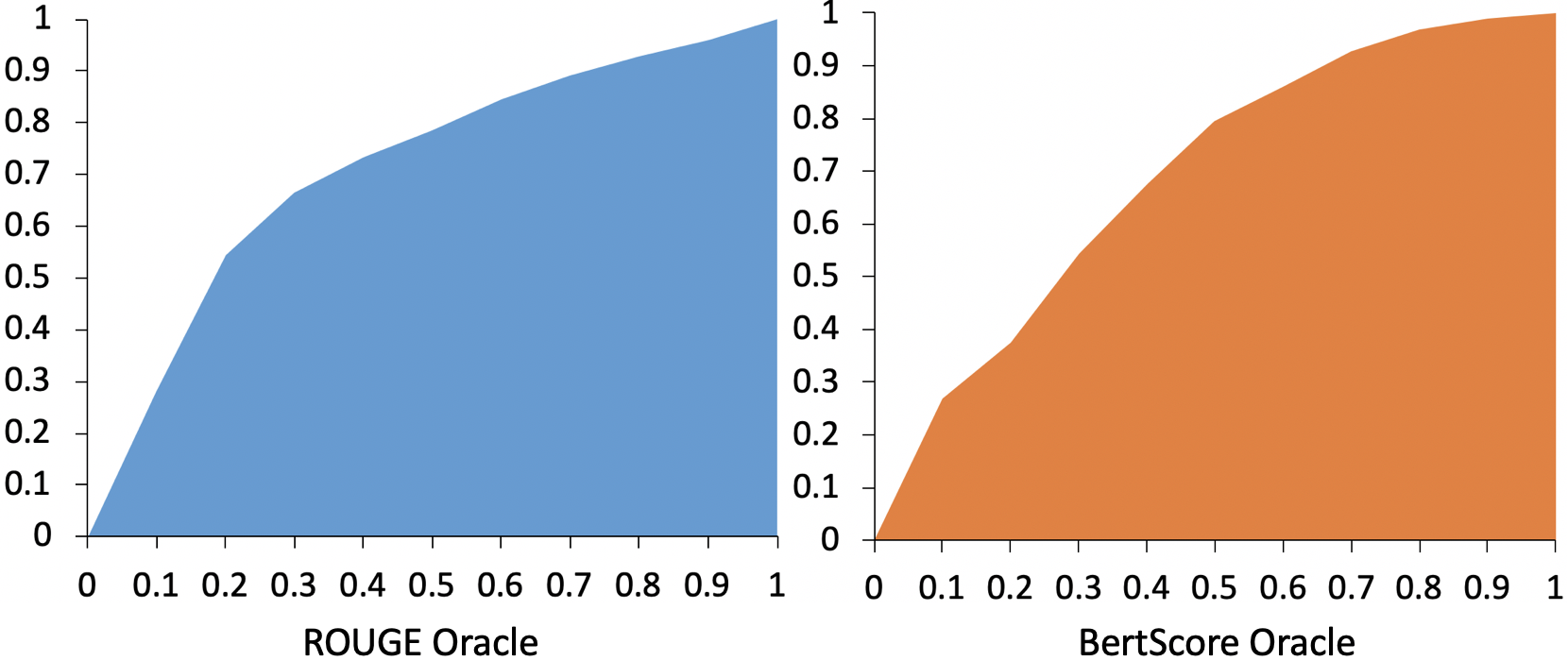}
\caption{Cumulative position distribution of oracles built on ROGUE (Blue) and BertScore (Orange). X axis is the ratio of article length. Y axis is the cumulative percentage of summary sentences.}
\label{ctrl-oracle-distribute-fig}
\vspace{-0.2cm}
\end{figure}

\subsection{Oracle Construction and Evaluation}
To build oracles with semantic similarity, we first segment sentences in source documents and human-written gold summaries\footnote{See details of the corpus in Appendix \ref{apdx:corpus_detail}.}. Then we convert the text to a semantically rich distributed vector space. For each sentence in a gold summary, we use BertScore to calculate its semantic similarity with candidates from the source content, then the sentence with the highest recall score is chosen. Candidates with a recall score lower than $0.5$ are excluded to streamline the selection process.

We observed that the oracle summaries generated through semantic similarity differ from those chosen from n-gram overlap. The positional distributions of two schemes are different, where early sentence bias is less significant for the BertScore scheme (see Figure \ref{ctrl-oracle-distribute-fig}). To further evaluate the effectiveness of this oracle construction approach, we conducted two assessments. ROUGE scores were computed with the gold summaries. Table \ref{ground-human-table} shows oracle summaries derived from BertScore are comparable though slightly lower than those from ROUGE, which is not unexpected given that BertScore is mismatched with the ROUGE metric. 
We also conducted two human evaluations. First, we ranked the candidate summary pairs of 50 news samples based on their similarity to human-written gold summaries \cite{narayan-2018-ExtremeSumm}. Four linguistic analyzers were asked to consider two aspects: informativeness and coherence \cite{radev-hovy-2002-introSumm}. The evaluation score represents the likelihood of a higher ranking, and is normalized to $[0, 1]$. Next, we adopted the question-answering paradigm  \cite{liu-lapata-2019-PreSumm} to evaluate 30 selected samples. For each sentence in the gold summary, questions were constructed based on key information such as events and named entities. Questions where the answer can only be obtained by comprehending the full summary were also included. Human annotators were asked to answer these questions given an oracle summary. The extractive summaries constructed with BertScore are significantly higher in all human evaluations (see Table~\ref{ground-human-table}).

\begin{table}[t!]
\begin{center}
\small
\begin{tabular}{lp{1.5cm}p{1.5cm}}
\hline & \bf ROUGE-1 & \bf ROUGE-2 \\
& \bf F1 Score & \bf F1 Score \\
\hline
ROUGE Oracle & 51.84 & 31.08 \\
BertScore Oracle & 50.56 & 29.41 \\
\hline
\end{tabular}
\begin{tabular}{lp{1.8cm}}
\hline 
\textbf{Similarity Evaluation} & \textbf{Score} \\
\hline
Gold Summaries & - \\
ROUGE Candidates & 0.70 \\
BertScore Candidates & 0.84 \\
\hline
\hline 
\textbf{QA Paradigm Evaluation} & \textbf{Accuracy} \\
\hline
\textbf{Entity and Event Questions:} \\
Gold Summaries & 0.95 \\
ROUGE Candidates & 0.54 \\
BertScore Candidates & 0.72 \\
\hline
\textbf{Extended Questions:} \\
Gold Summaries & 0.87 \\
ROUGE Candidates & 0.52 \\
BertScore Candidates & 0.70 \\
\hline
\end{tabular}
\end{center}
\vspace{-0.2cm}
\caption{\label{ground-human-table} ROUGE and Human evaluation scores of oracle summaries built on BertScore and ROUGE.}
\vspace{-0.2cm}
\end{table}

\section{Sub-Aspect Control Codes}
\label{sec:ctrl_code}
\subsection{Sub-Aspect Features in News Summarization}
Conditional generation often uses control codes as an auxiliary vector to adjust  pre-defined style features. Classic examples include sentiment polarity in  style transfer \cite{john-2019-disentangledTrans} or physical attributes (e.g. color) in image generation \cite{higgins-2017-betaVAE}. However, for summarization it is challenging to pinpoint such intuitive or well-defined features, as the writing style could vary according to genre, topic, or editor preference.

In this work, we adopt \textit{position}, \textit{importance} and \textit{diversity} as a set of sub-function features to characterize extractive news summarization  \cite{jung-etal-2019-earlier}. Considerations include: (1) ``inverted pyramid'' writing style is common in news articles, thus making layout or position a salient sub-aspect for summarization; (2) Importance sub-aspect indicates the assumption that repeatedly occurring content in the source document contains more important information; (3) Diversity sub-aspect suggests that selected salient sentences should maximize the semantic volume in a distributed semantic space \cite{Lin-2012-summAspects, yogatama-2015-defineDiversity}.

\subsection{Summary-Level Quantitative Analysis}
\label{ssec:sample-feature-analysis}
We apply two methods to evaluate the compatibility and effectiveness of the sub-aspects we choose for extractive news summarization. First, we conduct a quantitative analysis on the CNN/Daily Mail corpus, based on the assumption that the writing style variability of summaries can be characterized through different combinations of sub-aspects \cite{Lin-2012-summAspects}.

\begin{figure}[t]
\centering
\includegraphics[width=5.5cm]{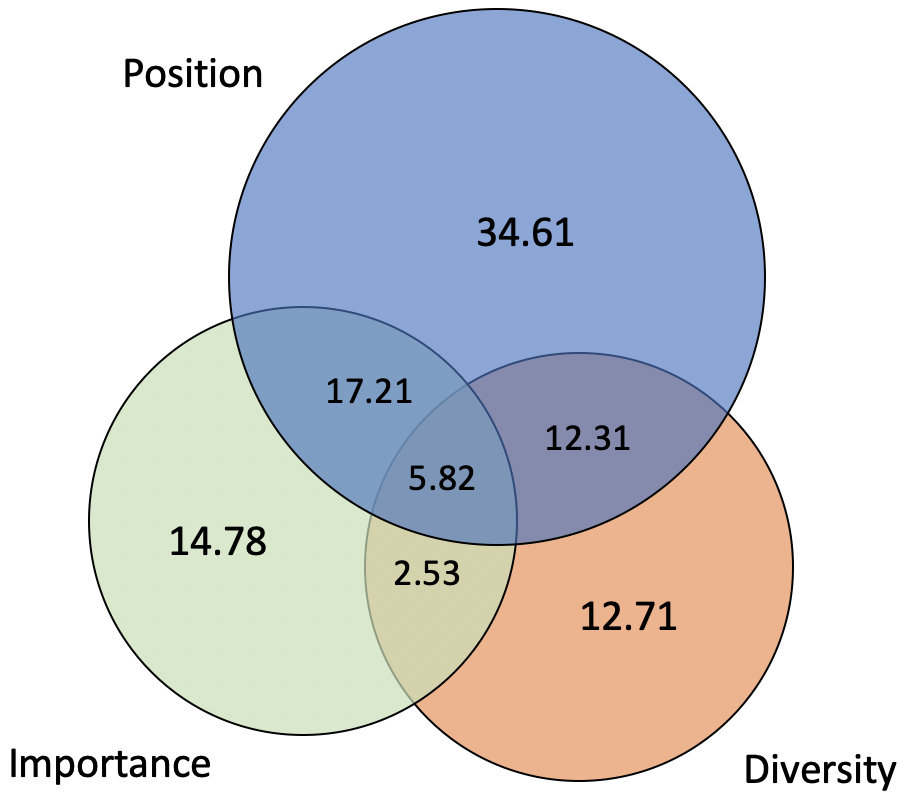}
\caption{Sample-level distribution of sub-aspect functions of the BertScore oracle. Values are the percentage in categorized samples, which adds up to 60.03\% of CNN/Daily Mail training set. The remaining 39.97\% do not belong to any of these 3 sub-aspects.}
\label{ctrl-feature-dist-fig}
\vspace{-0.3cm}
\end{figure}

For each source document, we converted all sentences to vector representations with a pre-trained contextual language model BERT \cite{devlin-2019-BERT}\footnote{https://github.com/google-research/bert}. For each sentence, we averaged hidden states of all tokens as the sentence embedding.
Similar to \cite{jung-etal-2019-earlier}, to obtain the subset of sentences which correspond to \textit{importance} sub-aspect, we adopted an N-Nearest method which calculates an averaged Pearson correlation between one sentence and the rest for all source sentence vectors, and collected the first-$k$ candidates with the highest scores ($k$ equals oracle summary length). To obtain the subset which corresponds to the \textit{diversity} sub-aspect, we used one implementation\footnote{http://www.qhull.org/} of the QuickHull algorithm \cite{barber-1996-quickhull} to find vertices, which can be regarded as sentences that maximize the volume size in a projected semantic space. For the subset that corresponds to the \textit{position} sub-aspect, the first 4 sentences in the source document were chosen.

With three sets of sub-aspects, we quantified the distribution of different sub-aspects on the extractive oracle constructed in Section \ref{sec:oracle-annotation}. An oracle summary will be mapped to the importance sub-aspect when at least two sentences in the summary are in the subset of \textit{importance} sub-aspect. For those oracle summaries that are shorter than 3 sentences (occupying 19\% of the oracle), only one sentence was used to determine which sub-aspect they would be mapped to. Note that the mapping is many to many; i.e. each summary can be mapped to more than one sub-aspect. Figure \ref{ctrl-feature-dist-fig} displays the distribution of the three sub-aspect functions of the oracle summaries, where \textit{position} occupies the largest area. This visualization shows that the three sub-aspects represent distinct linguistic attributes but could overlap with one another. 

\begin{figure}[t]
\centering
\includegraphics[width=7.3cm]{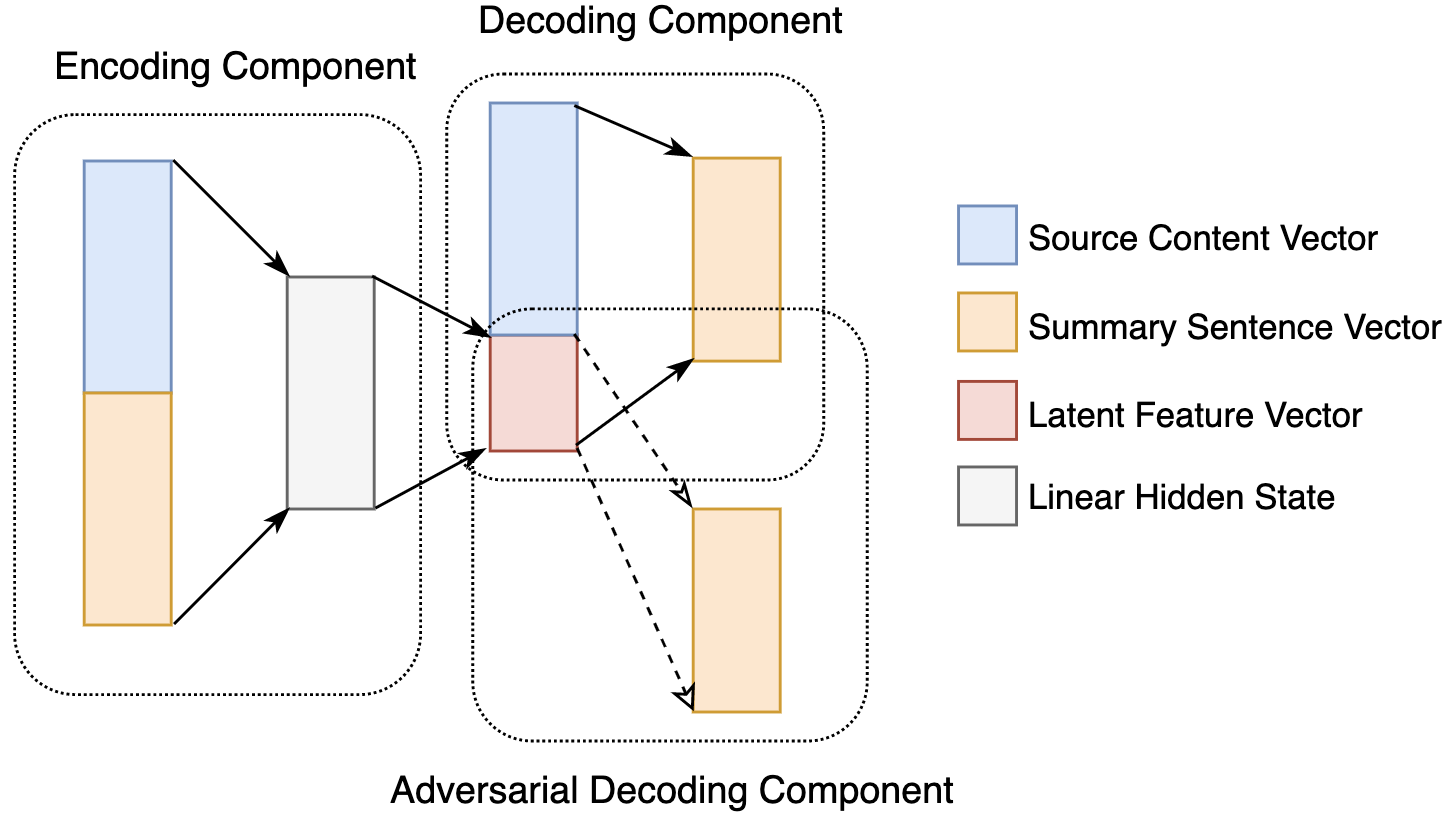}
\caption{Autoencoder with adversarial training strategy for unsupervised clustering of sentence-level distribution of sub-aspect functions.}
\label{ctrl-autoencoder-fig}
\vspace{-0.2cm}
\end{figure}

\begin{figure}[t]
\centering
\includegraphics[width=6.5cm]{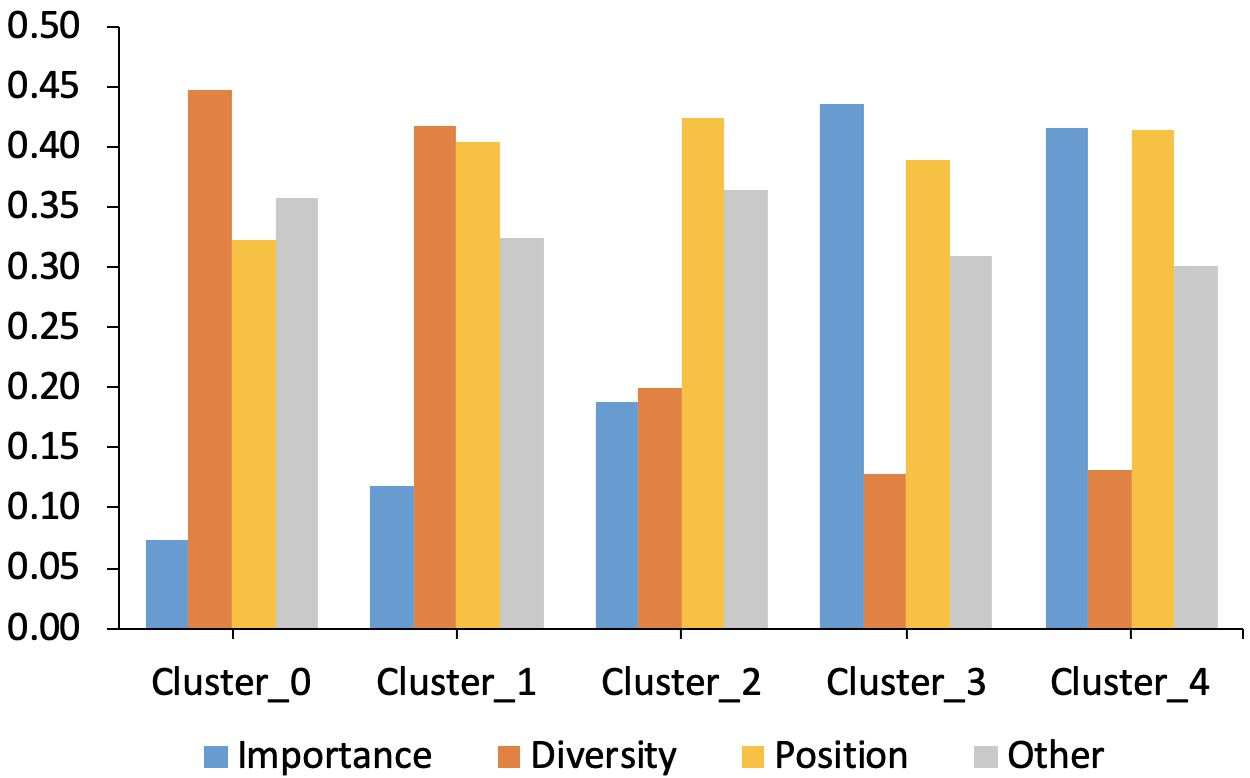}
\caption{Sentence-level clustering result labeled with sub-aspect features. X axis is the cluster index. Y axis is the proportion of sub-aspect features in each cluster.}
\label{ctrl-cluster-result-fig}
\vspace{-0.3cm}
\end{figure}

\subsection{Sentence-Level Unsupervised Analysis}
\label{ssec:unsupervised-analysis}
According to the mapping algorithm in the previous section, 39\% summaries were not mapped to a sub-aspect. This finding motivated us to investigate the distribution of sub-aspect functions \textit{at the sentence level}. Thus, we conducted unsupervised clustering, assuming that samples within one cluster are most similar to each other and they can be represented by the dominant feature.

As shown in Figure \ref{ctrl-autoencoder-fig}, we use an autoencoder architecture with adversarial training to model the correlation between document and summary sentences in the semantic space. The encoding component receives the source document representation and one summary sentence representation as input, and compresses it to a latent feature vector. Then, the latent vector and document vector are concatenated and fed to the decoding component to reconstruct the sentence vector. To obtain a compact yet effective latent vector representing the correlation between the source and summary, we adopt an adversarial training strategy as in \cite{john-2019-disentangledTrans}. More specifically, the adversarial decoder we include aims to reconstruct the sentence vector directly from the latent vector. During the training process, we update parameters of the autoencoder with an adversarial penalty (see Appendix \ref{apdx:autoencoder} for implementation details). 
After training this autoencoder, we conduct k-means clustering ($k=5$) on the latent representation vectors. Then, we analyze the clustering output, with the sentence-level labels of sub-aspect functions as defined in Section \ref{ssec:sample-feature-analysis}. As shown in Figure \ref{ctrl-cluster-result-fig}, sentences with position sub-aspect is distributed relatively equally across each cluster, while importance and diversity dominate in respectively different clusters. Based on the clustering results, we assign the sub-aspect function which is dominant to unmapped sentences in the same cluster. For instance, diversity is assigned to unmapped sentences in cluster 0 and 1 while importance is assigned to those in cluster 3 and 4. By doing this, we reduce $\approx 78\%$ of unmapped sentences and further reduce 35\% unmapped summaries using the same criteria in Section \ref{ssec:sample-feature-analysis}.

\section{Conditional Neural Generation}
\label{sec:experiment}
In this section, we construct a set of control codes to specify the three sub-aspect features described in Section \ref{sec:ctrl_code}, and label the oracle summaries constructed in Section \ref{sec:oracle-annotation}, then we propose a neural extractive model with a conditional learning strategy for a more flexible summary generation.

\subsection{Control Code Specification Scheme}
The control codes are constructed in the form of \textit{[importance, diversity, position]} to specify sub-aspect features. We can flexibly indicate the \textit{`ON'} and \textit{`OFF'} state of each sub-aspect by switching its corresponding value to $1$ or $0$, thus enabling disentanglement of each sub-aspect function. For instance, the control code $[1,0,0]$ would tell the model to focus more on importance during sentence scoring and selection, while $[0,1,1]$ would focus on both diversity and position. Indeed, switching the position code to $0$ would help the model obtain minimal position bias. Note that this does not mean the first few sentences would not be selected, as there is overlap between position, importance and diversity (shown in Figure \ref{ctrl-feature-dist-fig}). There are 8 control codes under this specification scheme, and we expect this code design can provide the model with sub-aspect conditions for generating summaries. 

\subsection{Neural Extractive Selector}
\label{ssec:neural_selector}
Given a document $D$ containing a number of sentences $[s_0, s_1, ... , s_n]$, the content selector assigns a score $y_i \in {[0,1]}$ to each sentence $i$, indicating its probability of being included in the summary. A neural model can be trained as an extractive selector for text summarization tasks by contextually modeling the source content. 

\begin{figure}[t]
\centering
\includegraphics[width=6.5cm]{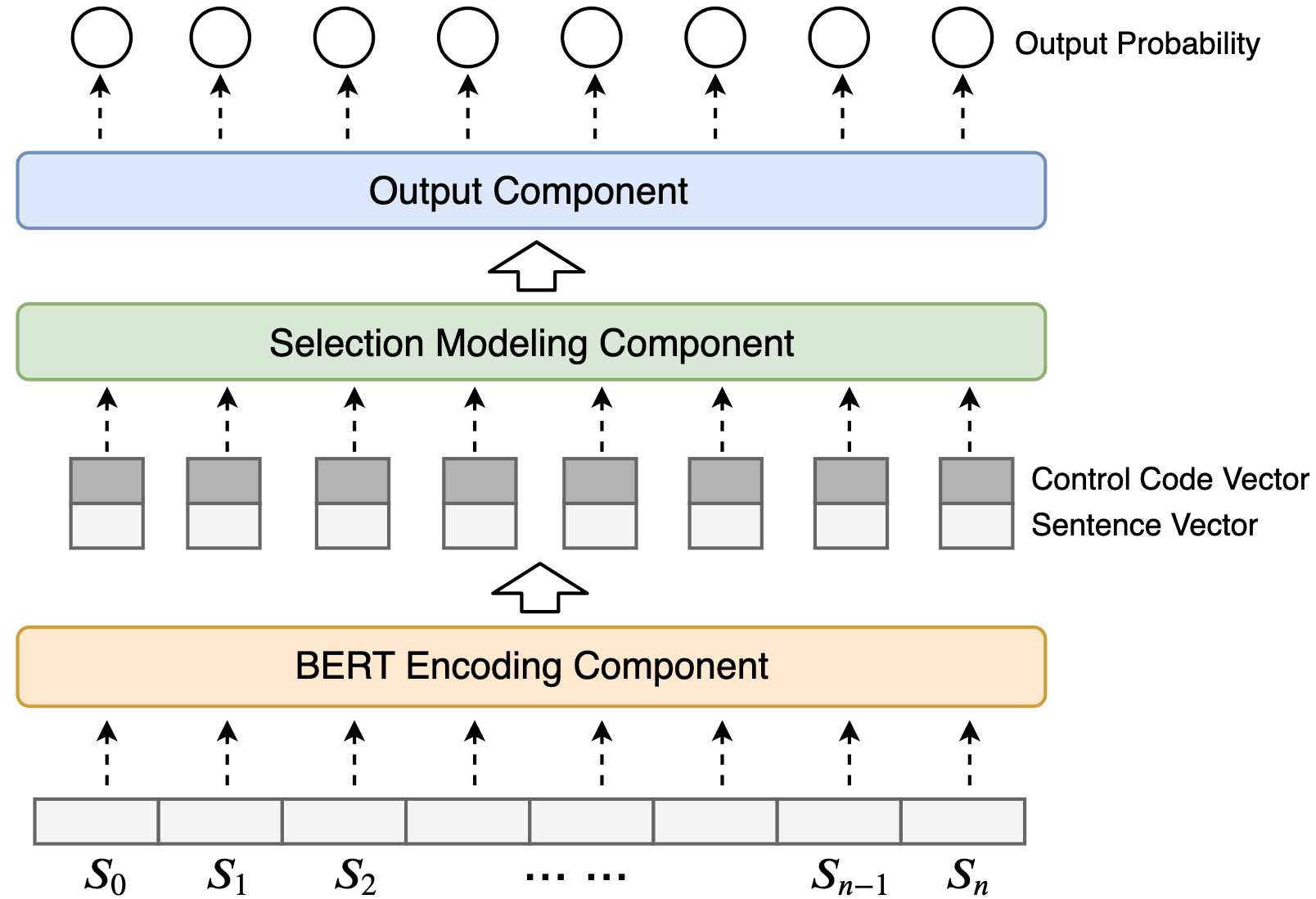}
\caption{Overview of the neural selector architecture.}
\label{ctrl-selector-fig}
\end{figure}

Here, we implemented and adapted the neural extractive selector in a sequence labeling manner \cite{kedzie-2018-contentSelection}. As shown in Figure \ref{ctrl-selector-fig}, the model consists of three components: a contextual encoding component, a selection modeling component and an output component. First, we used BERT in the contextual encoding component to obtain feature-rich sentence-level representations. Then, in the training process, we concatenated these sentence embeddings with the pre-calculated control code vector and fed them to the next layer, which models the contextual hidden states with the conditional signals. Next, a linear layer with Sigmoid function receives the hidden states and produces scores for each segment between 0 and 1 as the probability of extractive selection. While this architecture is straightforward, it has shown to be competitive when combined with state-of-the-art contextual representation \cite{liu-lapata-2019-PreSumm}.

In our setting, sentences were processed by a sub-word tokenizer \cite{wu2016wordpiece} and their embeddings were initialized with 768-dimension ``base-uncased'' BERT \cite{devlin-2019-BERT} and were fixed during training. Lengthy source documents were not truncated. For the selection modeling component, we applied a multi-layer Bi-directional LSTM \cite{schuster1997-biLSTM} and a Transformer network \cite{vaswani-2017-Transformer} and it was empirically shown that a two-layer Bi-LSTM performed best (see Appendix \ref{apdx:selector} for more implementation details). During testing, sentences with the top-3 selection probability were extracted as output summary, and we used the Trigram Blocking strategy \cite{paulus-2017-rl-summ} to reduce redundancy.

\begin{figure}[t]
\centering
\includegraphics[width=7.7cm]{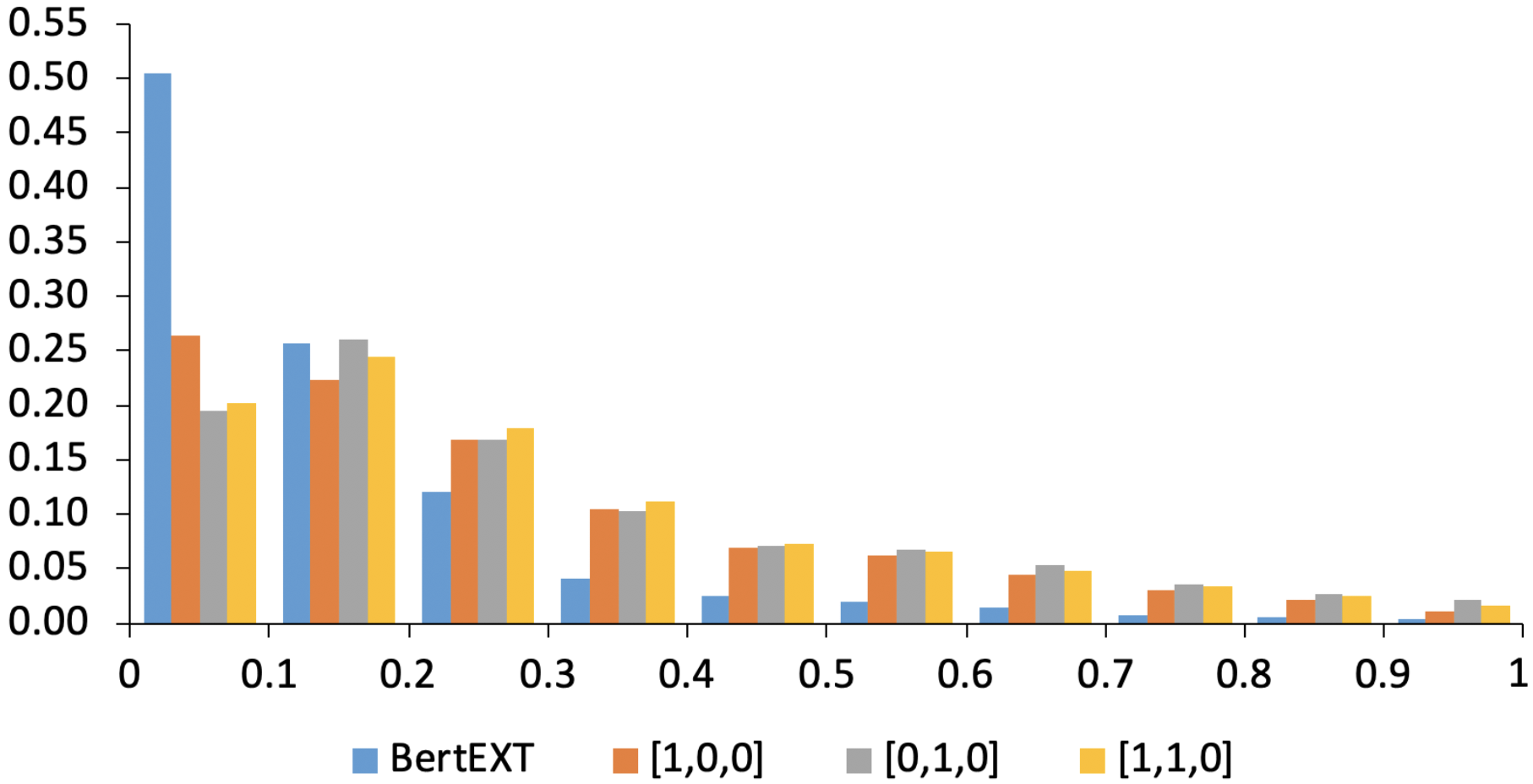}
\caption{Position distribution of generated summaries from a strong baseline model BertEXT and our conditional summarization model with position code set to $0$ (3 implementations). X axis is the position ratio. Y axis is the sentence-level proportion.}
\label{ctrl-codes-position-fig}
\vspace{-0.2cm}
\end{figure}

\section{Experimental Results and Analysis}
\label{sec:result_analysis}

\subsection{Quantitative Analysis}

To test the possibility of reducing position bias by conditioning summary generation, we switched the position code to $0$ and compared the position of selected sentences in summaries generated by our model to the state-of-the-art baseline BertEXT, based on fine-tuning BERT \cite{liu-lapata-2019-PreSumm}. The results show that BertEXT has a 50\% chance of choosing the first 10\% of sentences in the document. While the proposed framework still has a stronger tendency to choose sentences from the first 30\% of the sentences, its position distribution is flattened compared to that of BertEXT.

\begin{figure}[t]
\centering
\includegraphics[width=7cm]{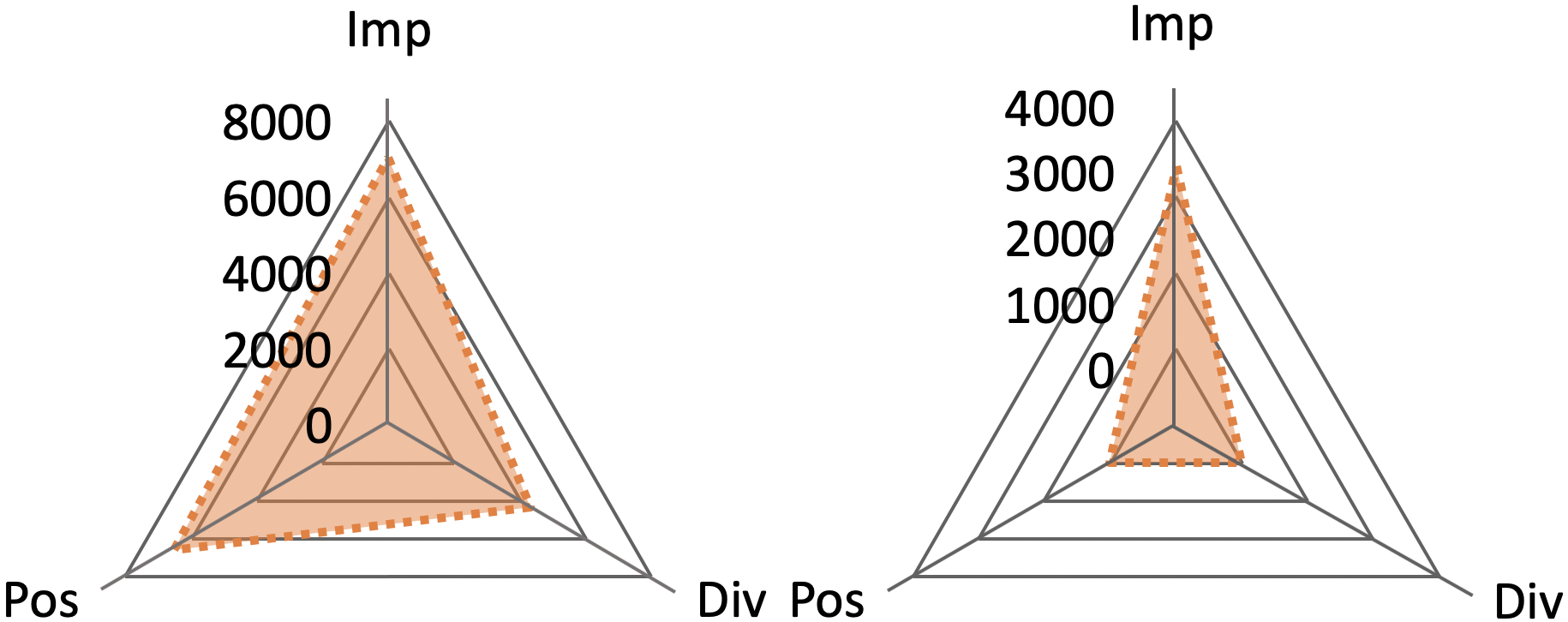}
\caption{Sub-aspect mapping of generated summary with importance-focus code [1,0,0]. Left panel: one sentence in the summary belongs to importance sub-aspect. Right panel: two sentences in the summary belong to importance sub-aspect. Contour lines denote the number of generated summaries.}
\label{ctrl-radar-imp-fig}
\vspace{-0.2cm}
\end{figure}

\begin{figure}[t!]
\centering
\includegraphics[width=7cm]{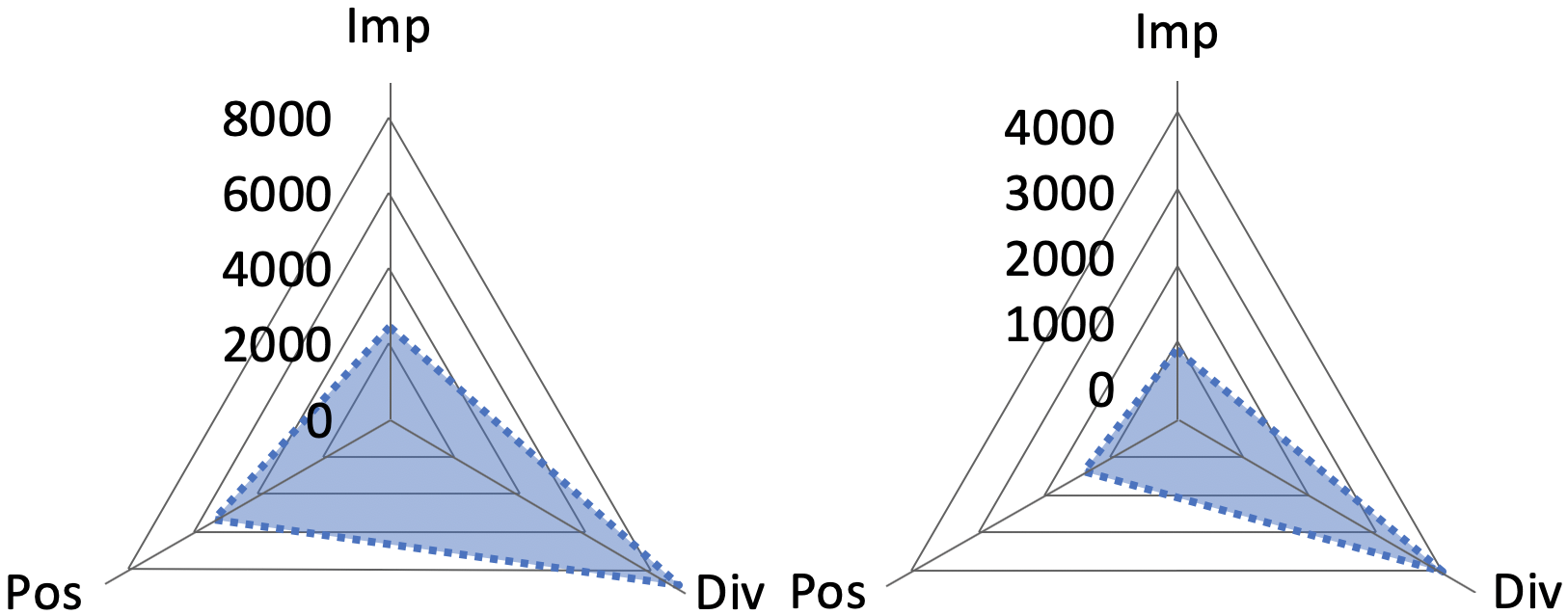}
\caption{Sub-aspect mapping of generated summary with diversity-focus code [0,1,0]. Left panel: one sentence in the summary belongs to diversity sub-aspect. Right panel: two sentences in the summary belong to diversity sub-aspect. Contour lines denote the number of generated summaries.}
\label{ctrl-radar-div-fig}
\end{figure}

We respectively switched importance and diversity codes to $1$ and categorized the generated summaries into subset of each sub-aspect function as in Section \ref{ssec:sample-feature-analysis}. As shown in Figure \ref{ctrl-radar-imp-fig} and \ref{ctrl-radar-div-fig}, summaries in the subset of importance and diversity weigh higher when the corresponding control codes are ON. Together, these results demonstrate the feasibility of our proposed framework, which can generate output summaries of alternative styles when given different control codes.

\subsection{Automatic Evaluation}
\label{ssec:automatic_eval_result}
We calculated F1 ROUGE scores for generated summaries under 8 control codes, and compared them with the BertScore oracle (see Section \ref{sec:oracle-annotation}), the Lead-3 baseline by selecting first-3 sentences as summary, and several competitive extractive models: SummaRuNNer \cite{nallapati2017summarunner}, TransformerEXT and BertEXT \cite{liu-lapata-2019-PreSumm}. From Table \ref{output-ROUGE-table} we observe that: (1) Summary generated from code [0,0,1] is similar to LEAD-3 but can dynamically learn the positional features not limited to the first 3 sentences, while isolating out diversity and importance features. (2) Only focusing on the importance sub-aspect leads to the worst performance, but performance can be improved when considering other sub-aspects. (3) Focusing on the diversity sub-aspect (i.e. Code [0,1,0]) can generate results comparable to strong baselines.

\begin{table}[t!]
\begin{center}
\small
\begin{tabular}{ccc}
\hline & \bf ROUGE-1 & \bf ROUGE-2 \\
\hline
 Oracle (BertScore) & 50.56 & 29.41 \\
 LEAD-3 & 40.42 & 17.62 \\
\hline
 SummaRuNNer* & 39.60 & 16.20 \\
 TransformerEXT* & 40.90 & 18.02 \\
 BertEXT* & 43.23 & 20.24 \\
\hline
 Code [0,0,0] & 39.44 & 17.37\\
 Code [0,0,1] & 40.21 & 18.25 \\
 Code [0,1,0] & 39.18 & 17.11 \\
 Code [0,1,1] & 40.70 & 18.42 \\
 Code [1,0,0] & 36.72 & 14.74 \\
 Code [1,0,1] & 40.33 & 17.90 \\
 Code [1,1,0] & 37.59 & 15.68 \\
 Code [1,1,1] & 40.87 & 18.50 \\
\hline
\end{tabular}
\end{center}
\caption{\label{output-ROUGE-table} ROUGE F1 score evaluation with various control codes, in the form of \textit{[importance, diversity, position]}. * denotes the results from corresponding paper.}
\vspace{-0.2cm}
\end{table}

\subsection{Human Evaluation} 
In addition to automatic evaluation, the human evaluation was conducted by experienced linguistic analysts using Best-Worst Scaling \cite{louviere2015bestRanking}. Analysts were given 50 news articles randomly chosen from the CNN/Daily Mail test set and the corresponding summaries from 6 systems: the oracle, BertEXT, three codes disabling sub-aspect position, and one code enabling position. They were asked to decide the best and the worst summaries for each document in terms of informativeness and coherence \cite{radev-hovy-2002-introSumm,narayan-2018-ExtremeSumm}. We collected judgments from 5 human evaluators for each comparison. For each evaluator, the documents were randomized differently. The order of summaries for each document was also shuffled differently for each evaluator. The score of a model was calculated as the percentage of times it was labeled as \textit{best} minus the percentage of times it was labeled as \textit{worst}, ranging from $-1.0$ to $1.0$. Since these differences come in pairs, the sum of all the evaluation scores for all summary types adds up to zero. We observed that summaries under diversity code are more favored than those under importance, and their combination can further produce better results (see Table \ref{output-human-table}). These findings resonate those from the automatic evaluation, suggesting that whether the evaluation metric is lexical overlap (ROUGE) or human judgement, the \textit{diversity} sub-aspect plays a more salient role than \textit{importance}. Moreover, both automatic and human evaluations show that summarizing with semantic-related sub-aspect condition codes achieves reasonable summaries. Examples in Appendix \ref{apdx:sample} show that generated summaries are not position-biased yet still preserve key information from the source content.

\begin{table}[t!]
\begin{center}
\small
\begin{tabular}{cc}
\hline & \bf Evaluation Score \\
\hline
 Oracle & 0.0458 \\
 BertEXT & 0.0332 \\
 Code [1,0,0] & -0.062 \\
 Code [0,1,0] & 0.0198 \\
 Code [0,0,1] & -0.071 \\
 Code [1,1,0] & 0.0350 \\
\hline
\end{tabular}
\end{center}
\caption{\label{output-human-table} Human evaluation on samples from baselines and our model with control codes, in the form of \textit{[importance, diversity, position]}.}
\vspace{-0.2cm}
\end{table}

\begin{table}[t!]
\begin{center}
\small
\begin{tabular}{ccc}
\hline & \bf ROUGE-1 & \bf ROUGE-2 \\
\hline
 BertEXT & 36.78 (-6.45) & 14.95 (-5.29) \\
 Code [1,0,0] & 33.94 (-2.78) & 13.04 (-1.70) \\
 Code [0,1,0] & 36.59 (-2.59) & 14.33 (-2.78) \\
 Code [0,0,1] & 30.34 (-9.87) & 8.90 (-9.35) \\
\hline
\end{tabular}
\end{center}
\caption{\label{shuffle-table} Inference scores on samples with shuffled sentences. Control codes are in the form of \textit{[importance, diversity, position]}. Values in brackets: absolute decrease from scores on original in-order samples.}
\vspace{-0.2cm}
\end{table}

\subsection{Inference on Samples of Shuffled Sentences} 
To further assess the decoupling between using sub-aspect signals and positional information learned by the model, we conducted an experiment on samples with shuffled sentences, similar to document shuffle in \cite{kedzie-2018-contentSelection}. In our setting, we only introduce the shuffle process in the model inference phase. We shuffled the sentences of all test samples we used in Section \ref{ssec:automatic_eval_result}, then applied the well-trained model to generate the predicted summaries. As shown in Table \ref{shuffle-table}, outputs under position sub-aspect and BertEXT suffer a significant drop in performance when we shuffle the sentence order. By comparison, there is far less decrease between the shuffled and in-order samples under diversity and importance control code, demonstrating that the latent features of these two semantic-related sub-aspects rely less on the position information, suggesting that applying semantic sub-aspects in the training process can reduce systemic bias learned by the model on a corpus with strong position preference.

\begin{table}[t!]
\begin{center}
\small
\begin{tabular}{cccc}
\hline & \bf R-1 F1 & \bf R-2 F1 & \bf R-2 Recall \\
\hline
 Oracle  & - & - & 8.70* \\
 Baseline  & - & - & 6.10* \\
 BertEXT & 26.91 & 3.70 & 2.98 \\
 Code [1,0,0] & 34.81 & 6.23 & 6.34 \\
 Code [0,1,0] & 31.79 & 5.32 & 4.62 \\
 Code [0,0,1] & 29.67 & 3.98 & 3.47 \\
\hline
\end{tabular}
\end{center}
\caption{\label{ami-result-table} Inference scores on AMI corpus from baselines and our model with control codes, in the form of \textit{[importance, diversity, position]}. * denotes results from \cite{kedzie-2018-contentSelection}.}
\vspace{-0.2cm}
\end{table}

\subsection{Inference on AMI Meeting Corpus} 
We also conducted an inference experiment on a less position-biased corpus. The AMI corpus \cite{carletta2005ami} is a collection of meetings annotated with text transcriptions with human-written summaries. Different from news summarization, meeting summaries are abstractive with extracted keywords. 
Unlike the previous comparison work in \cite{kedzie-2018-contentSelection}, we did not train the model from scratch with the AMI training set. Instead, we only applied the pre-trained model (without any fine-tuning) in Section \ref{sec:result_analysis} for summarization inference on its test set (20 meeting transcript-summary pairs).
Table \ref{ami-result-table} shows summaries under importance code obtain the highest ROUGE-1 and ROUGE-2 scores, better than the best-reported model in \cite{kedzie-2018-contentSelection}. Not surprisingly, summaries under the position code do not perform well, as there is less position bias in AMI. These findings suggest that our models with semantic-related control codes generalize across domains.

\section{Conclusion}
We proposed a neural framework for conditional extractive news summarization. In particular, sub-aspect functions of \textit{importance}, \textit{diversity} and \textit{position} are used to condition summary generation. This framework enables us to reduce position bias, a long-standing problem in news summarization, in generated summaries while preserving comparable performance with other standard models. Moreover, our results suggest that with conditional learning, summaries can be more efficiently tailored to different user preferences and application needs.

\section*{Acknowledgments}
This research was supported by funding from the Institute for Infocomm Research (I2R) under A*STAR ARES, Singapore. We thank Ai Ti Aw, Bin Chen, Shen Tat Goh, Ridong Jiang, Jung Jae Kim, Ee Ping Ong, and Zeng Zeng at I2R for insightful discussions.
We also thank the anonymous reviewers for their precious feedback to help improve and extend this piece of work.

\bibliography{emnlp2020}
\bibliographystyle{acl_natbib}

\clearpage

\appendix

\section{Details of the Corpus}
\label{apdx:corpus_detail}
The CNN/Daily Mail corpus \cite{hermann-2015-cnnDaily} contains English news articles and associated human-written summaries and is the most popular large-scale benchmark in news summarization. We used the pre-processed dataset as in \cite{see-2017-pgnet}, which has 287,226 training pairs, 13,368 validation pairs, and 11,490 test pairs. We did not replace the name entities with anonymised identifiers, and used the same sentence segmentation for documents and summaries as in \cite{liu-lapata-2019-PreSumm}. To obtain the word embedding representation, we tokenized the sentences with the sub-word algorithm from BERT \cite{devlin-2019-BERT}.

\section{Implementation Details of Unsupervised Analysis Model}
\label{apdx:autoencoder}
In this section, we provide implementation details of the model in Section \ref{ssec:unsupervised-analysis}: an autoencoder with adversarial training strategy. 

\noindent{\textbf{Encoding Component:}}
Given a document representation vector $\mathbf{v}_{doc}$, and a sentence representation vector $\mathbf{v}_{sen}$ as input, the encoding component (two linear layers) compresses it to a lower dimension, namely the latent feature vector $\mathbf{v}_{latent}$. In our setting, the hidden dimensions of $\mathbf{v}_{doc}$, $\mathbf{v}_{sen}$ and $\mathbf{v}_{latent}$ are 768, 768 and 10, respectively. $\mathbf{h}_{enc}$ is the hidden vector, defined as: 
\begin{equation*}
    \mathbf{h}_{enc} = \mathrm{LeakyRelu}(\mathbf{W}[\mathbf{v}_{doc};\mathbf{v}_{sen}]+\mathbf{b}) 
\end{equation*}

\noindent the lantent feature vector is defined as: 

\begin{equation*}
    \mathbf{v}_{latent} = \mathrm{Sigmoid}(\mathbf{W}\mathbf{h}_{enc}+\mathbf{b})
\end{equation*}
\noindent where $\mathbf{W}$ and $\mathbf{b}$ are trainable parameters in each layer, and ; denotes the concatenation operation.

\noindent{\textbf{Decoding Component:}}
Given a latent feature representation vector $\mathbf{v}_{latent}$ and a document representation $\mathbf{v}_{doc}$ as input, the decoding component (two linear layers) is targeted to reconstruct the sentence representation $\mathbf{v}_{sen}$.
\begin{equation*}
    \mathbf{h}_{dec} = \mathrm{LeakyRelu}(\mathbf{W}[\mathbf{v}_{doc};\mathbf{v}_{latent}]+\mathbf{b})
\end{equation*}
\begin{equation*}
    \mathbf{s}_{dec} = \mathbf{W}\mathbf{h}_{dec}+\mathbf{b}
\end{equation*}
\noindent where $\mathbf{h}_{dec}$ and $\mathbf{s}_{dec}$ are the hidden state and reconstruction output, respectively.

\noindent{\textbf{Adversarial Decoding Component:}}
Given a latent feature representation vector $\mathbf{v}_{latent}$ as input, the adversarial decoding component (one linear layer) is targeted to reconstruct the sentence representation $\mathbf{v}_{sen}$.

\begin{equation*}
    \mathbf{s}_{adv} = \mathbf{W}\mathbf{v}_{latent}+\mathbf{b}
\end{equation*}
\noindent where $\mathbf{s}_{adv}$ is the reconstruction output.

\noindent{\textbf{Training Procedure and Setting:}}
During each training batch, there is a two-step parameter update:\\
1) Update the adversarial decoder with Mean Square Error (MSE) loss between $\mathbf{s}_{adv}$ and $\mathbf{v}_{sen}$.
\begin{equation*}
    loss_{adv} = \mathrm{MSE}(\mathbf{s}_{adv}, \mathbf{v}_{sen})
\end{equation*}
2) Update the autoencoder with MSE loss between $\mathbf{s}_{adv}$ and $\mathbf{v}_{sen}$, combined with a penalty from the adversarial MSE to reduce the unnecessary information leaked from $\mathbf{v}_{sen}$ in the encoding component. The adversarial loss is defined as:
\begin{equation*}
    loss_{adv} = \mathrm{MSE}(\mathbf{s}_{dec}, \mathbf{v}_{sen}) - \lambda \mathrm{MSE}(\mathbf{s}_{adv}, \mathbf{v}_{sen}) 
\end{equation*}
\noindent where $\lambda=0.2$ in our training setting.

Adam optimizer \cite{kingma2014adam} was used with learning rate of $1e{-}3$ and weight decay of $1e{-}3$. Batch size was set to 64. Drop-out \cite{srivastava2014dropout} of $rate=0.1$ was applied in each linear layer. BERT parameters were fixed during training. The trainable parameter size was 0.6M. Tesla V100 with 16G memory was used for training and we used the Pytorch 1.4.0 as computational framework\footnote{https://github.com/pytorch/pytorch}.

\section{Implementation Details of Neural Selector Model}
\label{apdx:selector}
In this section, we provide implementation details of the model in Section \ref{ssec:neural_selector}: a neural sentence selector for extractive summarization.

\noindent{\textbf{BERT Encoding Component:}}
Given a document $\mathbf{D}$ containing a number of sentences $[\mathbf{s}_0, \mathbf{s}_1, ... , \mathbf{s}_n]$ as input, the encoding component produces the sentence representation $\mathbf{h}_i$ from each $\mathbf{s}$, which is a list of tokens $[\mathbf{w}_0, \mathbf{w}_1, ..., \mathbf{w}_m]$. Here we use the average of the token-level hidden states in the last layer of BERT as $\mathbf{h}_i$.

\begin{equation*}
    \mathbf{h}_{i} = \frac{1}{m}\sum^m_i \mathbf{w}^{BertRep}_i
\end{equation*}

\noindent{\textbf{Selection Modeling Component:}}
Given the specific control code $\mathbf{v}_{ctrl}$ and sentence vectors $\mathbf{H}=[\mathbf{h}_0, \mathbf{h}_1, ... ,\mathbf{h}_n]$ as input, this component use a two-layer bi-directional LSTM to model the contextual information with sub-aspect conditioning. The forward and backward hidden states are concatenated as output.
\begin{equation*}
    \mathbf{u}^{forward}_{i} = \mathrm{LSTM}_{f}([\mathbf{h}_{i-1};\mathbf{v}_{ctrl}])
\end{equation*}
\begin{equation*}
    \mathbf{u}^{backward}_{i} = \mathrm{LSTM}_{b}([\mathbf{h}_{i+1};\mathbf{v}_{ctrl}])
\end{equation*}
\begin{equation*}
    \mathbf{u}_{i} = [\mathbf{u}^{forward}_{i}; \mathbf{u}^{backward}_{i}]
\end{equation*}
\noindent where the input embedding dimension, hidden size, and control code dimension is 768, 384 and 3 respectively. ; denotes the concatenation operation.

\noindent{\textbf{Output Component:}}
A linear layer is used to produce output $\mathbf{y}_i$ for each sentence, as the probability of being included in the generated summary.
\begin{equation*}
    \mathbf{y}_{i} = \mathrm{Sigmoid}(\mathbf{W}\mathbf{u}_i + \mathbf{b})
\end{equation*}

\noindent{\textbf{Training Setting:}}
Binary cross entropy (BCE) is used to measure the loss between the prediction $y_i$ and the ground-truth $\hat{y}_{i}$ for all time steps:
\begin{equation*}
loss = \sum \mathrm{BCELoss}(\mathbf{\hat{y}}_{i}, \mathbf{y}_i).
\end{equation*}

Adam optimizer \cite{kingma2014adam} was used with learning rate of $3e{-}4$ and weight decay of $1e{-}4$. Batch size was set to 64. Drop-out \cite{srivastava2014dropout} of $rate=0.2$ was applied in the modeling layer and output linear layer. BERT parameters were fixed during training. Lengthy documents were not truncated. The trainable parameter size was 14M (excluding the pre-trained language model). We trained the model for 20 epochs (about 12 hours) on the Tesla V100 GPU. The reported models were selected based on the best validation performance, according to the inflection point of loss value.

\section{Generated Summary Examples}
\label{apdx:sample}
\begin{figure*}[ht]
\centering
\includegraphics[width=13.5cm]{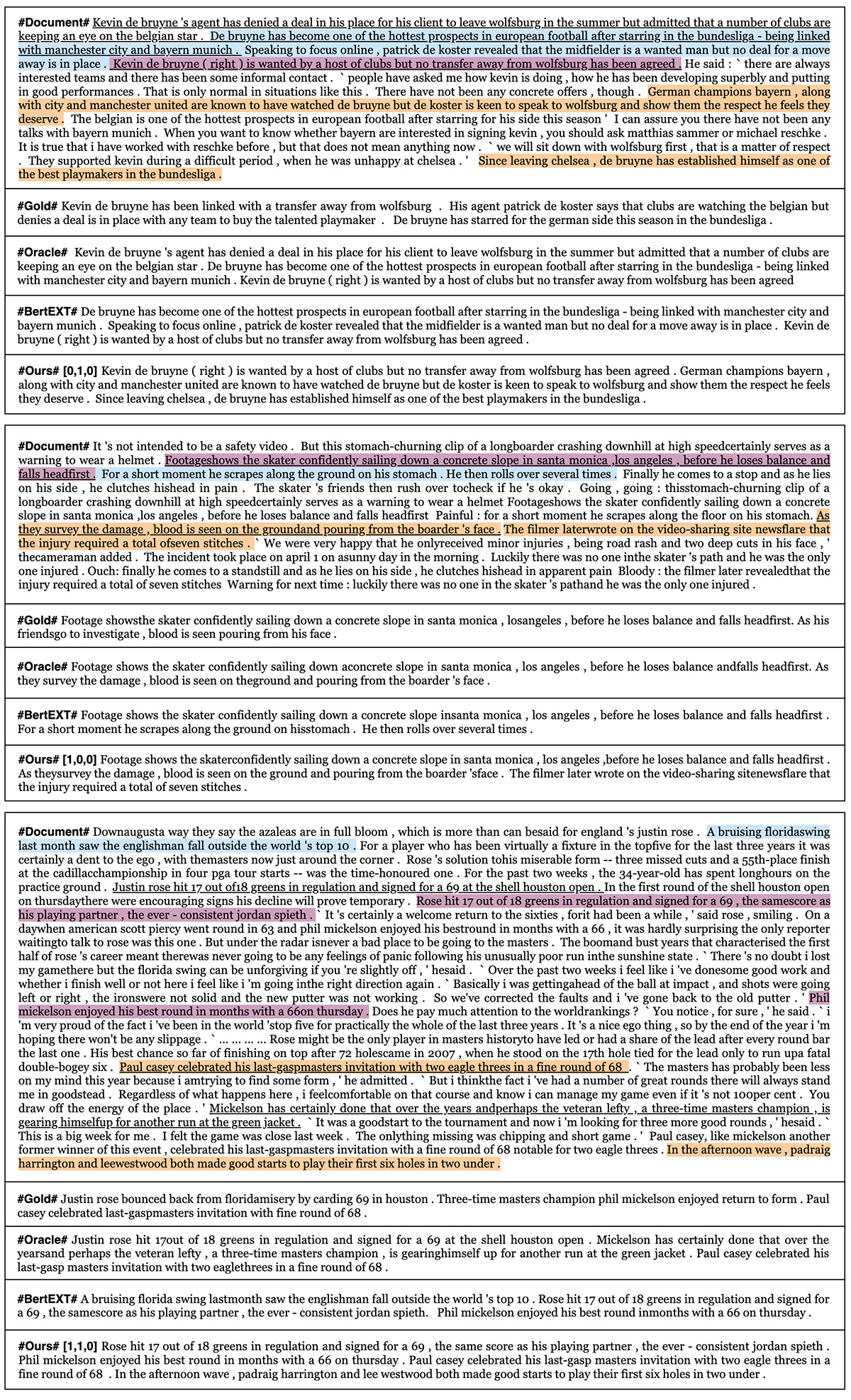}
\caption{Three news article examples. Oracle summaries are underlined, summaries from a baseline model are highlighted in blue, summaries from our model with specified control codes are in orange, and their overlaps are in purple.}
\label{ctrl-apx-example-fig}
\vspace{-1cm}
\end{figure*}

We show summary examples from the golden groundtruth, oracle, the baseline BertEXT, and the proposed conditional neural summarizer for three news articles in Figure \ref{ctrl-apx-example-fig}.
\vspace{-1cm}


\end{document}